%% file: gpl_iclr2022_conference.tex
\documentclass{article} 
\usepackage{gpl_iclr2022_conference,times}

\input{math_commands.tex}

\usepackage{hyperref}
\usepackage{url}
\usepackage{graphicx}

\title{Imitation Learning for Generalizable Self-driving Policy with Sim-to-real Transfer}

\author{Zoltán Lőrincz \& Márton Szemenyei \\
Department of Control Engineering and Information Technology \\ Budapest University of Technology and Economics \\
Budapest, 1117, Hungary \\
\texttt{zoltan.lorincz@edu.bme.hu, szemenyei@iit.bme.hu} \\
\And
Róbert Moni \\
Department of Telecommunication and Media Informatics\\
Budapest University of Technology and Economics\\
Budapest, 1117, Hungary \\
\texttt{robertmoni@tmit.bme.hu}
}

\iclrfinalcopy 
\begin{document}

\maketitle

\begin{abstract}
Imitation Learning uses the demonstrations of an expert to uncover the optimal policy and it is suitable for real-world robotics tasks as well. 
In this case, however, the training of the agent is carried out in a simulation 
environment due to safety, economic and time constraints. Later, the agent is applied in 
the real-life domain using sim-to-real methods. In this paper, we apply Imitation 
Learning methods that solve a robotics task in a simulated environment and use transfer 
learning to apply these solutions in the real-world environment. Our task is set in the 
Duckietown environment, where the robotic agent has to follow the right lane based on the 
input images of a single forward-facing camera. We present three Imitation Learning and 
two sim-to-real methods capable of achieving this task. A detailed comparison is provided 
on these techniques to highlight their advantages and disadvantages.
\end{abstract}

\section{Introduction}

Imitation Learning (IL) uses demonstrations of an expert to uncover the optimal policy. 
Due to this, the agent can achieve expert-like behavior in the given environment.

IL is a feasible approach for problems where collecting labeled data is complicated, 
however acquiring expert demonstrations is a straightforward process. One such area 
is robotic control, as it is usually challenging to solve the particular task with a 
rule-based policy, but collecting demonstrations is in most cases is uncomplicated. 
As a result, IL has been widely used in this area. 

ALVINN \citep{NIPS1988_812b4ba2} was one of the first imitation learning based 
self-driving solutions. It used Behavioural Cloning (BC) to carry out a real-world 
lane following task. \citet{DBLP:journals/corr/abs-1011-0686} presented improvements 
to the Behavioural Cloning formula and solved a self-driving task in a 3D racing 
game. \citet{DBLP:journals/corr/LiSE17} extended the Generative Adversarial Imitation 
Learning algorithm to learn a policy that can distinguish certain behaviors in human 
driving in a racing simulator.

Despite all of these, solving a real-world robotics task using solely Imitation Learning 
is problematic. The preferred approach is to train a model in a simulator and deploy it 
in the real-world domain. However, this is challenging as the model's performance usually 
declines in the real-world due to the differences of the two environments. To address this 
issue, it is recommended to apply sim-to-real techniques. The aim of these methods is to 
augment the learning process in a way that an algorithm trained in one domain would achieve
similar performance in a different domain. 

One such technique is Domain Randomization. Instead of training the model in a single 
simulated environment, different parameters of the simulator are randomized to expose the 
model to a wide range of environments at training time. With enough variability, the real 
world may appear to the model as just another variation of the simulator. This way the model 
will learn general features that are applicable to the real world as well. The randomized 
variables of the simulator are usually either visual parameters (e.g. textures, lighting 
conditions, etc.) or physical parameters (e.g. friction coefficients, the gravitational 
acceleration, masses, sizes or other attributes of objects, etc.).

Several works have demonstrated that Domain Randomization can successfully solve the
simulator-to-real problem. \citet{DBLP:journals/corr/TobinFRSZA17} uses this method for 
object localization on real-world images by training a neural network in a simulator with 
highly manipulated images. \citet{DBLP:journals/corr/abs-1710-06537} uses randomization 
of the simulator’s dynamics to train a neural network which moves objects to the assigned 
locations using a robotic arm. \citet{DBLP:journals/corr/abs-1910-07113} uses Automatic 
Domain Randomization to train a robotic arm to solve a Rubik’s cube. This technique 
incrementally increases the applied Domain Randomization and thus the difficulty of the 
environment, as the model learns to perform well in the previous environments. By performing 
the task in more and more difficult conditions, the model learns to generalize. As a result, 
the model trained in the simulator can successfully work on the physical, real-world robot.

In case of vision-based algorithms, a feasible way to perform the domain transfer is by 
applying Visual Domain Adaptation. The aim of this technique is to transfer the observations 
from the training and testing domains to a common domain, which is then used to train the 
agent to perform the given task. 

Due to recent advances in image-to-image translation, this approach is becoming more and 
more popular. \citet{DBLP:journals/corr/abs-1812-03823} uses Visual Domain Adaptation to 
train a self-driving agent that achieves equally good performance in both the simulated 
and real-world domains. Their model uses an Unsupervised Image-to-Image Translation Network 
\citep{DBLP:journals/corr/LiuBK17} to translate images between the real-world operating 
domain and the generated simulation environment, while also learns to predict control 
decisions from the ground truth labels from the simulator.

In this work, several experiments were carried out in the Duckietown \citep{7989179} 
simulator environment to solve the presented lane following task using Imitation Learning. 
After the agent's performance was satisfactory in the simulator, we applied different 
sim-to-real techniques to bridge the gap between the simulation and the real environment.

\section{Proposed Methods}
\label{prop_methods}

In this work, we implemented and evaluated three different Imitation Learning 
techniques for the self-driving task of right-lane following in the Duckietown 
simulator environment \citep{gym_duckietown}. We chose the best performing 
method and applied two sim-to-real methods to solve the sim-to-real 
transfer problem.

\begin{figure}[h]
    \begin{center}
        \includegraphics[width=0.56\linewidth]{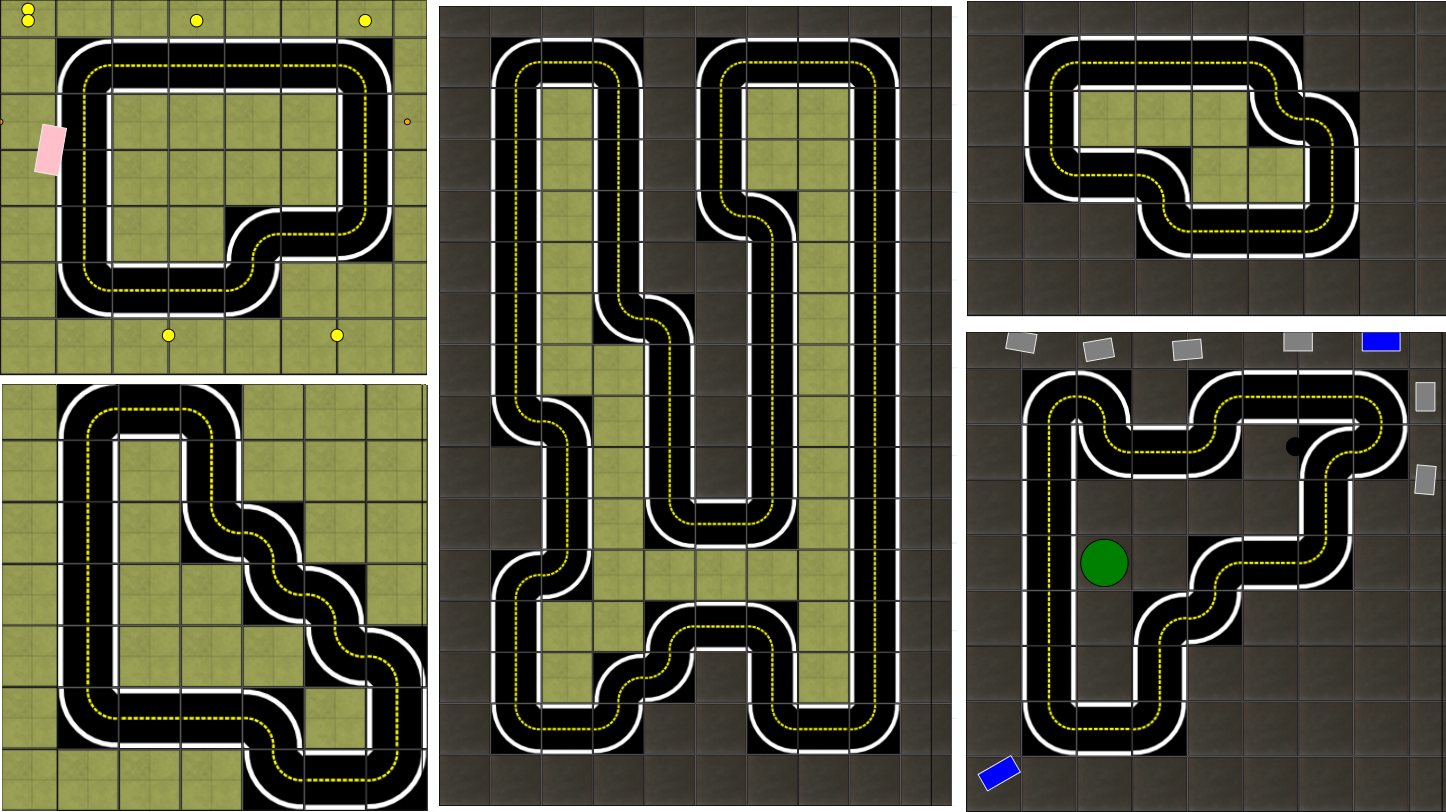}
    \end{center}
    \caption{Training maps}
    \label{figure-maps}
\end{figure}

\subsection{Imitation Learning in the simulation}

The general training procedure of the IL experiments was the following. First, the 
expert demonstrator was rolled out in the environment to collect training data - 
demonstrations. Next, the agent was trained based on the IL algorithm. Finally, the 
agent was released in the environment and evaluated using the official Duckietown 
metrics.

The demonstrations were collected on multiple maps. When the expert completed its 
current trajectory, the environment was reset and a new map was randomly selected 
from the available set of maps (see Figure~\ref{figure-maps}). Most of the maps 
contain several objects on the side of the roads to increase variability. This 
multi-map training approach further improves the model’s generalization ability and 
provides robustness on unseen track layouts.

\subsubsection{Applied Imitation Learning algorithms}

During our work we experimented with three IL algorithms: Behavioral Cloning (BC)
\citep{10.5555/647636.733043}, Dataset Aggregation (DAgger) 
\citep{DBLP:journals/corr/abs-1011-0686} and Generative Adversarial Imitation 
Learning (GAIL) \citep{DBLP:journals/corr/HoE16}.

\textbf{Behavioral Cloning:} is the simplest form of IL. It focuses on learning the 
expert’s policy using Supervised Learning. Expert demonstrations are divided into 
state-action pairs, these pairs are treated as independent and identically distributed 
examples and finally, Supervised Learning is applied.

\textbf{DAgger:} this method assumes, that there is access to an 
interactive demonstrator at training time. The algorithm starts with the initial 
predictor policy that had been uncovered from the initial expert demonstrations 
using Supervised Learning. Then, the following loop is executed until the algorithm 
converges. In each iteration, trajectories are collected by rolling out the current 
policy that had been obtained in the previous iteration and the state distribution 
is estimated. For every state feedback is collected from the expert and using this, 
a new policy is trained. For the algorithm to work efficiently, it is important to 
use all the previous training data during the teaching, so that the agent remembers 
all the errors it made in the past. Therefore, DAgger trains the actual policy on 
all the previous training data.

\textbf{GAIL:} is an Inverse Reinforcement Learning (IRL) algorithm. It aims to 
uncover a reward function by the means of the demonstrations, which is then used 
to learn the policy using Reinforcement Learning. GAIL adopts the Generative 
Adversarial Networks (GAN) \citep{goodfellow2014generative} architecture to carry 
out IRL. Similarly to GANs, the GAIL architecture consists of two neural networks: 
the policy network (or the generator) and the discriminator. The policy network 
acts as the agent’s policy: it receives the agent’s state in the environment as an 
input and outputs the adequate actions. The discriminator is a binary classifier 
which tries to distinguish the received state-action pairs from the trajectories 
generated by the agent and the expert. This network can be interpreted as the cost 
function that provides the learning signal to the policy.

\subsubsection{Expert demonstrator}

\begin{figure}[h]
    \begin{center}
        \includegraphics[width=0.9\linewidth]{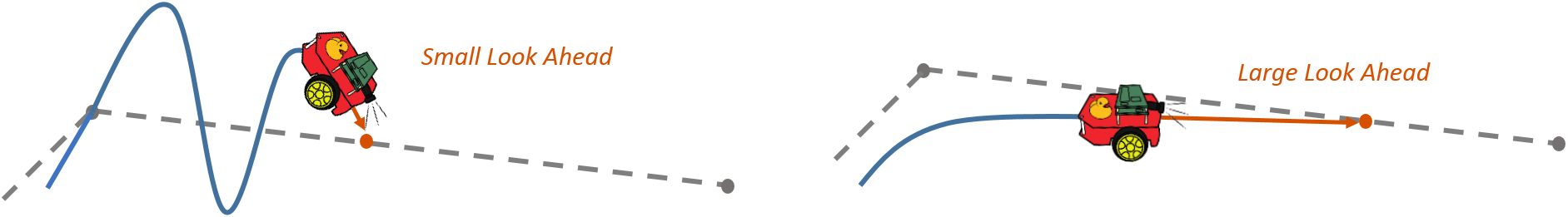}
    \end{center}
    \caption{The operation of the pure pursuit controller.}
    \label{figure-expert}
\end{figure}

The Duckietown software stack contains an implementation a pure pursuit controller. 
The algorithm uses the Duckiebot’s relative position and orientation to the center 
of the right driving lane to calculate the adequate actions of Pulse Width 
Modulation (PWM) signals. It selects a point on the ideal driving line at a certain 
distance from the agent and controls the robot to move towards this point. This is 
demonstrated by Figure~\ref{figure-expert}. 

Furthermore, we modified the pure pursuit controller to use different velocity and 
steering gain values for straights and for corners. We also extended the original
proportional controller with a derivative gain, which managed to further improve the 
controller’s performance.

We manually fine-tuned the pure pursuit PD controller and used it as the expert 
demonstrator in our Imitation Learning experiments, as this algorithm greatly 
outperforms a possible human demonstrator (controlling the robot with a joystick or 
a keyboard).

\subsubsection{Simplifying observations and actions}
The demonstrations are sequences of state-action pairs. In the case of the Duckietown 
simulator, the states are observations of the environment: images from the Duckiebot’s 
front camera; and the actions are PWM signals that specify the Duckiebot’s left and 
right motor velocities. To achieve better performance at training and inference, we simplified both the observations 
and the actions during experiments by applying a preprocessing step to the images 
and a postprocessing step to the actions.

Observations taken from the Duckietown simulator or from the Duckiebot are RGB images 
with the resolution of $\displaystyle 480$x$\displaystyle 640$ (height × width). Images 
of this size introduce a few problems to the learning algorithms. The high-resolution 
results in a high-dimensional state-space, which makes it harder for the algorithm to 
learn a proper feature extractor. It also slows down the inference and training time of 
the neural network. Therefore, before feeding the images to the models, two preprocessing 
steps are performed in order to reduce image complexity and increase training and inference 
speed. These are the following:

\begin{itemize}
\item\textit{Downscaling:} The images are resized to a smaller resolution of 
$\displaystyle 60$×$\displaystyle 80$ to reduce the dimensionality of the state-space 
and increase training speed.

\item \textit{Normalization:} The pixel values are converted to floating-point numbers 
and are normalized to the $\displaystyle [0.0, 1.0]$ range. This is a commonly used data 
preprocessing method that helps the training process by alleviating numerical problems 
of the optimization.
\end{itemize}

In case of the GAIL algorithm, these preprocessing steps are followed by feeding the 
preprocessed image through feature extractor: a ResNet \citep{DBLP:journals/corr/HeZRS15} 
network that was pretrained on the ImageNet \citep{5206848} dataset.

In both simulation and the real Duckietown environments, the Duckiebots are 
controlled by actions of PWM signals, which represent the left and right motor velocities. 
However, during experiments the models are trained to predict two actions: throttle 
and steering angle. The throttle action is a scalar value between $\displaystyle 0.0$ 
and $\displaystyle 1.0$, where $\displaystyle 0.0$ causes the agent to stop and in case 
of $\displaystyle 1.0$ the agent moves at full speed. The steering angle action is a 
scalar value between $\displaystyle -1.0$ and $\displaystyle 1.0$, where 
$\displaystyle -1.0$ and $\displaystyle 1.0$ causes the agent to turn fully to the left 
and right respectively, and in case of $\displaystyle 0.0$ the agent moves in a straight 
line. The actions predicted by the networks are then converted to PWM signals to suit 
the Duckiebots.

\subsection{Sim-to-real methods}

The agents trained in the simulation failed to perform the lane following task in the 
real Duckietown environment, as they could not generalize to the different, previously 
unseen real-world environment. Therefore, it was necessary to apply sim-to-real 
techniques to bridge the gap between the two environments. The aim of these 
techniques is to augment the learning process in a way that an algorithm trained in 
one visual domain would achieve similar performance in a different visual domain. We 
used two methods to solve the sim-to-real problem: Domain Randomization (DR) and Visual 
Domain Adaptation using Unsupervised Image-to-Image Translation Networks 
\citep{DBLP:journals/corr/LiuBK17} (VDA-UNIT).

\begin{figure}[h]
    \begin{center}
        \includegraphics[width=0.99\linewidth]{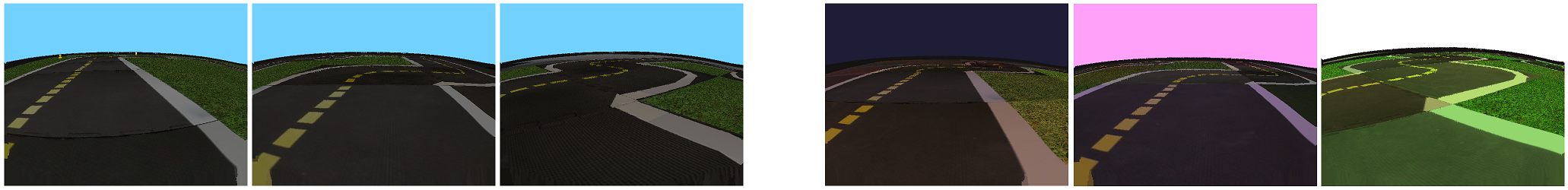}
    \end{center}
    \caption{Observations from the standard (left) and the domain randomized (right) environment.}
    \label{figure-domainrand}
\end{figure}

The Duckietown simulator has a built-in Domain Randomization functionality, which 
changes the parameters of the simulation (e.g. lighting conditions, textures, camera 
parameters, size of the robot, physical parameters, etc.) each time the simulation is 
reset (see Figure~\ref{figure-domainrand}). We applied this technique during the 
process of collecting demonstrations, so that later the agent would be trained on 
domain randomized observations.

\begin{figure}[h]
    \begin{center}
        \includegraphics[width=0.48\linewidth]{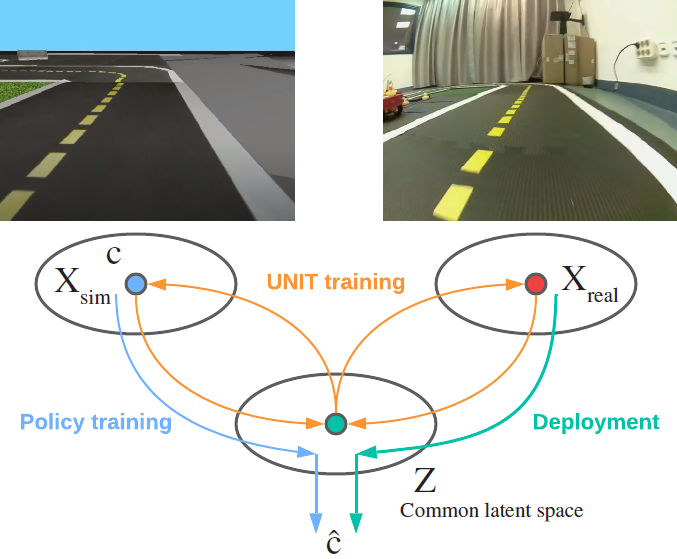}
    \end{center}
    \caption{UNIT network-based Visual Domain Adaptation.}
    \label{figure-unit}
\end{figure}

Our second solution adapts a UNIT network to transfer the observations from the 
simulated $\displaystyle \emX_{sim}$ and the real $\displaystyle \emX_{real}$ domains 
into a common latent space $\displaystyle \emZ$. After the UNIT network is properly 
trained and the quality of the image-to-image translation is satisfactory, the control 
policy is trained from this common latent space $\displaystyle \emZ$ using the 
demonstrations $\displaystyle \vc$ from the expert in the simulation. The method is 
demonstrated by Figure~\ref{figure-unit}.

The main advantage of this method is that it does not require pairwise correspondences
between images in the simulated and real-world training sets to perform the 
image-to-image translation. Furthermore, it does not require real-world labels either, 
the lane-following agent can be trained by using only the demonstrations from the 
simulation.

\section{Training and Evaluation setup}

\subsection{Training procedure}
\label{training-procedure}

We have carried out the following IL experiments: BC with DR, BC without DR, DAgger 
with DR, DAgger without DR and GAIL with BC-based pretraining. We also trained the 
Duckietown software stacks’ DAgger algorithm as a baseline solution.

For the BC experiments and the GAIL pretraining phase $\displaystyle 98304$ 
demonstrations were collected ($\displaystyle 128$ episodes × $\displaystyle 768$ 
timesteps). Throughout the DAgger experiments, the agent was rolled out for 
additional $\displaystyle 128$ episodes, for $\displaystyle 512$ timesteps per 
each episode. The acquired $\displaystyle 65536$ demonstrations were annotated 
by the expert and combined with the initial  demonstrations, which resulted in 
$\displaystyle 163840$  training examples. The collected demonstrations were 
randomly shuffled and split into training and validation datasets using 80\% 
and 20\% of the data. 

The training of the BC and the DAgger algorithms was performed using early stopping 
with patience set to $\displaystyle 25$ epochs. 

The training of the GAIL method started by pretraining the policy network using BC. 
After this, the entire GAIL algorithm was trained for $\displaystyle 30$ epochs. 
In each epoch, the agent was rolled out in the environment for $\displaystyle 15$ 
times, each trajectory consisted of $\displaystyle 256$ timesteps. The replay buffer 
could store $\displaystyle 75$ trajectories from the agents, which is 
$\displaystyle 19200$ observation-action pairs.

The models were trained with the Adam \citep{DBLP:journals/corr/KingmaB14} optimizer 
with the learning rate set to $\displaystyle 0.0001$. The batch size was set to 
$\displaystyle 32$. The Duckietown DAgger baseline was trained for $\displaystyle 50$ 
epochs with the default parameters.

The training of the VDA-UNIT sim-to-real method was conducted in two steps. 

As the first step, we trained the UNIT network, which was responsible for the 
image-to-image translation between the simulated and real images. For each environment, 
$\displaystyle 1024$ images were randomly sampled from the datasets of over 
$\displaystyle 30000$ images each. In the case of the real domain, the images were 
extracted from video feeds of real robots, while the simulated images were simply 
generated by running an agent in the simulation. The test datasets were set up 
similarly, by randomly sampling $\displaystyle 256$ images from each dataset 
(excluding those images that were already sampled for the training sets). Next, the 
model was trained for $\displaystyle 200$ epochs. We used the Adam optimizer with 
the learning rate set to $\displaystyle 0.0001$. After $\displaystyle 100$ epochs, 
a linear learning rate decay was applied. The batch size was set to $\displaystyle 4$.

As the second step, we selected the best performing IL algorithm (DAgger) as the 
controller's policy and we trained it using the  training procedure described earlier.

All of the experiments were run on a single NVIDIA GeForce RTX 2060 GPU.

\subsection{Evaluation procedure}
\label{eval-procedure}

To evaluate our algorithms in the simulation, we used the official Duckietown metrics. 
For the evaluation in the real-world environment, we defined custom metrics that could 
be easily measured. The metrics are presented in sections~\ref{metrics-sim} 
and~\ref{metrics-real} respectively.

The Duckietown software environment provides an evaluation interface, which deploys 
the given submission in the simulation, measures its performance by calculating the 
official performance metrics and creates a final report that contains all the results. 
The evaluation procedure runs the submission for 5 episodes in the environment, 
which means that the robot starts from a random position and operates for a fixed 
amount of time. The median values of the metrics are calculated from the results 
of these 5 runs. We used the official evaluation tool to evaluate our models in the 
simulation.

To evaluate the real-world algorithms the following procedure was used. For each model, 
two episodes were run, during which the robot was started from once in the inner and 
once in the outer loop. Each episode lasted for maximum of 60 seconds. If the robot 
left the track, the episode was terminated. In each episode the custom metrics were 
measured. Finally, the metrics during the 2 episodes were averaged.

\subsubsection{Performance metrics in the simulation}
\label{metrics-sim}

To evaluate our IL models in the simulation, we have used the four official Duckietown 
metrics. These are the following:
\begin{itemize}
\item \textit{Traveled distance:} This is the median distance traveled, along a lane. 
(That is, going in circles will not make this metric increase.) This is discretized 
to tiles. This metric only measures the distance that was travelled continuously 
(without cease) in the right driving lane. This metric encourages both faster driving 
as well as algorithms with lower latency.

\item \textit{Survival time:} This is the median survival time. The simulation is 
terminated when the robot goes outside of the road or it crashes with an obstacle.

\item \textit{Lateral deviation:} This is the median lateral deviation from the 
center line. This objective encourages “comfortable” driving solutions by penalizing 
large angular deviations from the forward lane direction to achieve smoother driving.

\item \textit{Major infractions:} This is the median of the time spent outside of 
the drivable zones. This objective means to penalize “illegal” driving behavior, for 
example driving in the wrong lane.
\end{itemize}

\subsubsection{Performance metrics in the real-world environment}
\label{metrics-real}

During the AI Driving Olympics \citep{DBLP:journals/corr/abs-1903-02503}
competitions, it is possible for the organizers to calculate the official Duckietown 
metrics, as the Duckietown tracks, where the submissions are evaluated, are equipped 
with complex positioning systems consisting of cameras, markers and precise computer 
vision algorithms. However, without access to the AIDO real-world evaluation system, 
these metrics are impossible to calculate, as the Duckiebots are not equipped with any 
positioning system or sensor. Therefore, we defined two custom real-world performance 
metrics that can be feasibly measured without such system:

\begin{itemize}

\item \textit{Survival time:} The time until the robot left the track or the time 
of the evaluation procedure (if the robot did not make a mistake).

\item \textit{Visited road tiles (in the correct driving lane):} The number of 
visited road tiles during the evaluation procedure. Only those tiles are counted, 
where the robot traveled inside in the right driving lane. Measuring the traveled 
distance of the robot is problematic, but counting the tiles instead is quite 
simple, therefore this is a feasible alternative.
\end{itemize}

\subsubsection{Baseline algorithms}
\label{baseline-algos}

The Duckietown software stack contains different baseline solutions for the 
challenges \footnote{\url{https://docs.duckietown.org/daffy/AIDO/out/embodied_strategies.html}}. One of the Imitation Learning baselines is a DAgger algorithm, which 
has a training procedure that is similar to our implementation. We trained this 
model with the default parameters, based on the instructions that were provided 
in the authors’ description. We used the resulted model as a baseline to measure 
and compare the performance of our algorithms in the simulation.

The baseline for the real-world experiments was a model trained using the best 
performing IL algorithm (DAgger) without applying any of the sim-to-real techniques.

\section{Results}

\subsection{Results in the simulation environment}

The models were evaluated with the Duckietown evaluator tool using the AIDO 
performance metrics. Table~\ref{results-sim} presents the best results for each 
training algorithm.

\begin{table}[h!]
    \caption{Evaluation in the simulation}
    \label{results-sim}
    \begin{center}
       \resizebox{0.7\textwidth}{!}{%
       \begin{tabular}{lccccc}
  
            \textbf{IL method} & \textbf{Survival Time} & \textbf{Traveled Distance}  & \textbf{ Lateral deviation} & \textbf{ Major Infractions} \\
            \hline
            \hline
            \textbf{BC} & & & \\
            w/ DR  &15     &5.26   &0.65   &0.23\\
            w/o DR &15     &5.44   &0.75   &0.63\\
            \hline
            \textbf{DAgger} & & & & \\
            w/ DR  &15     &5.32   &0.71   &\textbf{0}\\
            w/o DR &15     &\textbf{5.67}   &0.63   &\textbf{0}\\
            
            \hline
            \textbf{GAIL} & & & & \\
            w/ DR   &13.55  &4.78   &0.71   &1.27\\
            \hline
            \textit{BASELINE} &15   &3.97  &\textbf{0.35}   &\textbf{0}\\

        \end{tabular}}
    \end{center}
\end{table}

All three algorithms managed to train a reasonably well performing model. The agents 
were able to follow the right driving lane, without committing any crucial mistakes 
such as leaving the road. 

The algorithms managed to outperform the baseline model in terms of the traveled 
distance and survival time (except GAIL). The baseline, however, has a significantly 
lower lateral deviation. This is due to the fact that the baseline agent moves a lot 
slower than the trained agents.

As we can see, the GAIL algorithm has a slightly worse performance compared to DAgger 
and BC. The reason for this phenomenon might be complexity of the training procedure: 
the parameters of the training process are probably not well chosen. Therefore, further 
optimization is needed for the GAIL algorithm.

\subsection{Results in the real environment}
The models were evaluated based on the procedure described in 
section~\ref{eval-procedure} using the custom real-world metrics. 
Table~\ref{results-real} presents the best results for each training algorithm. It is 
important to note, that the robot’s driving speed was fine-tuned for each algorithm 
in order to maximize the survival time metric.

\begin{table}[h!]
    \caption{Sim-to-real experiments}
    \label{results-real}
    \begin{center}
        \resizebox{0.7\textwidth}{!}{%
        \begin{tabular}{lccc}
            \textbf{Sim-to-Real method} & \textbf{Average survival time} & \textbf{Average visited road tiles} \\
            \hline
            \hline
            DR                  &60     &20\\
            VDA-UNIT                &60     &18.5\\
            \textit{DAgger w/o sim2real}   &\textit{1}      &\textit{4.5}\\
        \end{tabular}}
    \end{center}
\end{table}

Both methods managed to successfully solve the sim-to-real problem, as the 
real-world robots could properly follow the right driving lane, without committing 
crucial mistakes. It is also straightforward, that in the real environment these 
techniques are not only useful, but necessary, as the model without any form of 
Transfer Learning completely failed at the lane-following task.

\subsubsection{The quality of the image-to-image translation}
\begin{figure}[h]
    \begin{center}
        \includegraphics[width=0.65\linewidth]{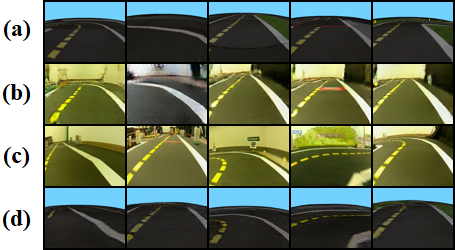}
    \end{center}
    \caption{a) simulation b) sim-to-real translation c) real d) real-to-sim translation}
    \label{figure-image-to-image}
\end{figure}

As it can be observed on Figure~\ref{figure-image-to-image}, the UNIT network achieves 
high image-to-image translation quality. The network managed to learn how to remove the 
background that is above the horizon and replace it with the blue sky when performing 
the real-to-sim translation. This is also true for the other way around: during the 
sim-to-real translation, the network removes the sky and replaces it with background 
objects.

\section{Conclusion and Future Work}

In this paper, we used Imitation Learning techniques to solve a complex self-driving 
robotics task in the Duckietown environment. We trained the models in the simulator 
and applied sim-to-real methods to ensure that the algorithms achieve equally good 
performance in the real-world environment. We evaluated the performance of the models 
in both environments using the Duckietown metrics. We showed that using our approach, 
trained agents were able to follow the right driving lane in both the simulated and 
real-world domains.

Our results demonstrate that it is favorable to use DAgger as it achieves the best 
performance with slightly more training time compared to BC. It is challenging to reach 
good performance with GAIL, as the training times are fairly longer and the hyperparameters 
are difficult to fine-tune.

In the future, we would like to continue fine-tuning the presented solutions in a 
hope of achieving even better results. Performing further optimizations on the GAIL 
algorithm should also be advantageous, as this model was the one with the poorest 
performance. In addition to this, we plan to apply Curriculum Learning to solve the 
more complex Duckietown challenges.

The source code of our work is available on GitHub\footnote{\url{https://github.com/lzoltan35/duckietown_imitation_learning}}.

\subsubsection*{Acknowledgments}
The research presented in this work has been supported by Continental Automotive Hungary Ltd.

\bibliography{gpl_iclr2022_conference}
\bibliographystyle{gpl_iclr2022_conference}


\end{document}

%% file: math_commands.tex
\usepackage{amsmath,amsfonts,bm}

\def\eqref#1{equation~\ref{#1}}

\def\1{\bm{1}}

\def\vc{{\bm{c}}}

\DeclareMathAlphabet{\mathsfit}{\encodingdefault}{\sfdefault}{m}{sl}
\SetMathAlphabet{\mathsfit}{bold}{\encodingdefault}{\sfdefault}{bx}{n}

\def\emX{{X}}

\def\emZ{{Z}}

